%% file: acl_latex.tex
\pdfoutput=1

\PassOptionsToPackage{dvipsnames}{xcolor}

\documentclass[11pt]{article}
\usepackage{arydshln}

\usepackage[]{acl}

\usepackage{hyperref}

\usepackage{times}
\usepackage{latexsym}
\usepackage{multirow}
\usepackage{booktabs}
\usepackage{hyperref}
\usepackage{CJK}
\usepackage{amsmath}
\usepackage{microtype}

\usepackage{arydshln}
\usepackage{tcolorbox} 
\usepackage[dvipsnames]{xcolor}
\tcbuselibrary{skins}
\usepackage{multirow}
\usepackage[normalem]{ulem}
\useunder{\uline}{\ul}{}
\usepackage{inconsolata}

\usepackage[T1]{fontenc}
\usepackage{epsfig}

\usepackage[utf8]{inputenc}
\usepackage[misc,geometry]{ifsym}

\newtcolorbox{prompt}[1]{
    colback=white,
    colframe=BlueViolet,
    fonttitle=\bfseries\color{white},
    colbacktitle=BlueViolet,
    title=#1
}

\usepackage{microtype}

\title{PerSphere: A Comprehensive Framework for Multi-Faceted Perspective Retrieval and Summarization }

\author{%
Yun Luo$^{1}$, 
Yingjie Li$^{1}$, 
Xiangkun Hu$^{2}$\thanks{~~The work does not relate to the author's position at Amazon.},
Qinglin Qi$^{3}$, \\
\textbf{Fang Guo}$^{1}$,
\textbf{Qipeng Guo}$^{4}$,
\textbf{Zheng Zhang}$^{5}$,
\textbf{Yue Zhang}$^{1,6 \ \textrm{\small\Letter}}$\thanks{~~Corresponding author.} \\
\centerline{\normalfont{$^1$School of Engineering, Westlake University} \quad \normalfont{$^2$Amazon AWS AI.}} \\
\centerline{ \quad \normalfont{$^3$Sichuan University.}\quad \normalfont{$^4$Shanghai AI Lab.}\quad \normalfont{$^5$New York University Shanghai.}} \\
\centerline{\normalfont{$^6$ Institute of Advanced Technology, Westlake Institute for Advanced Study}} \\
\centerline{\texttt{\{luoyun, zhangyue\}@westlake.edu.cn}}
}

\begin{document}
\maketitle
\begin{abstract}
As online platforms and recommendation algorithms evolve, people are increasingly trapped in echo chambers, leading to biased understandings of various issues. To combat this issue, we have introduced PerSphere, a benchmark designed to facilitate multi-faceted perspective retrieval and summarization, thus breaking free from these information silos. For each query within PerSphere, there are two opposing claims, each supported by distinct, non-overlapping perspectives drawn from one or more documents. Our goal is to accurately summarize these documents, aligning the summaries with the respective claims and their underlying perspectives. This task is structured as a two-step end-to-end pipeline that includes comprehensive document retrieval and multi-faceted summarization. Furthermore, we propose a set of metrics to evaluate the comprehensiveness of the retrieval and summarization content. Experimental results on various counterparts for the pipeline show that recent models struggle with such a complex task. Analysis shows that the main challenge lies in long context and perspective extraction, and we propose a simple but effective multi-agent summarization system, offering a promising solution to enhance performance on PerSphere.
\end{abstract}

\input{intro}

\input{dataset}

\input{method}

\input{results}

\section{Conclusion}
We proposed a new task for multi-faceted perspective retrieval and summarization, which addresses the limitation of the existing work \cite{chen-etal-2019-seeing} that requires perspective pools in advance or overlooking of the comprehensiveness, by adopting a RAG pipeline. For evaluation, we construct a dataset named PerSphere, and created a set of specific evaluation metrics tailored to this task.  Performance of various LLMs on PerShpere highlighted the significant challenges inherent in such a complex task. Motivated by a thorough analysis, we devised a straightforward yet powerful multi-agent summarization method, HierSphere, designed to enhance model performance on PerSphere. The effectiveness of this method has been confirmed through experiments.

\section*{Limitations}
In the task of argument mining \cite{ajjour-etal-2019-modeling,friedman-etal-2021-overview,Kamalloo}, the volume of data is roughly comparable to ours. In real-world contexts involving large data sets, we could start with a coarse retrieval using BM25 on smaller subsets (like 10,000 or 5,000 documents), and then proceed to re-rank them. For our study, since the goal is to evaluate the comprehensiveness of retrieval models and the summarization abilities of LLMs, our dataset is adequate without needing to perform coarse retrieval from millions of documents. But we could further enhance the size of the data by adding irrelevant documents such as cnn\_dailymail \cite{see-etal-2017-get}.
In terms of retrieval, although it poses a challenge for current models, addressing this issue is not the focus of our study and could be considered in future work.

\section*{Ethical Consideration}
We honor the ACL Code of Ethics. We collected data for Theperspective by automatically extracting data from www.theperspective.com. The CEO of the website, Daniel Ravner, granted us permission to extract
and use their data for academic research.   All annotators have received labor fees corresponding to the amount of their annotated instances. 



\bibliography{custom}
\bibliographystyle{acl_natbib}

\clearpage
\appendix

\input{appendix}

\end{document}

%% file: intro.tex
\section{Introduction}

\begin{figure}[t]
    \centering
    \includegraphics[width=0.9\hsize]{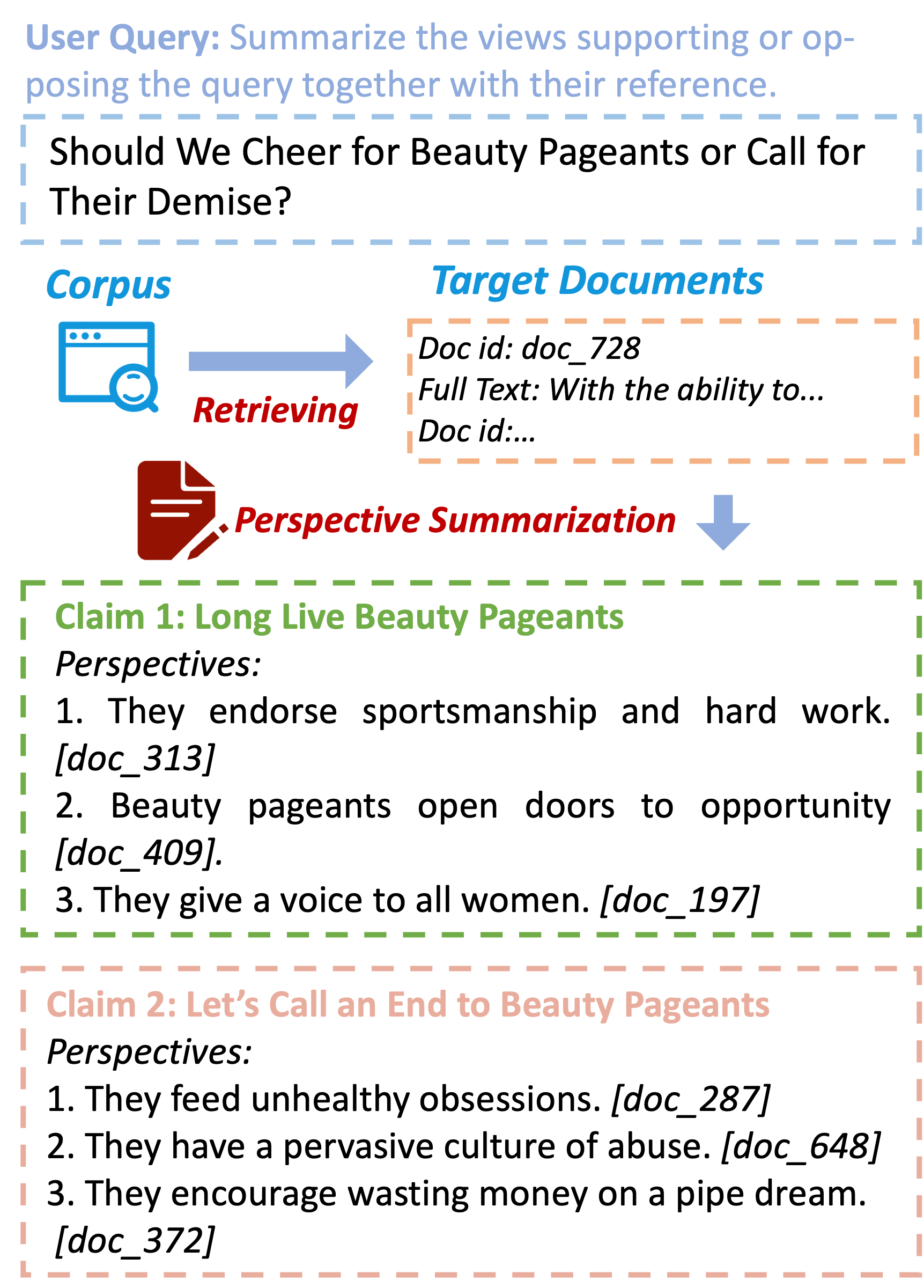}
    \caption{Multi-faceted perspective retrieval and summarization, where the system outputs a summary of two controversial claims and several non-overlapping perspectives together with corresponding  references.
    }
    \label{exp}
        \vspace{-3mm}
\end{figure}

The rapid expansion of social platforms and personalized recommendation systems on the web has significantly increased the risk of confining individuals within echo chambers \cite{nguyen2020echo}, limiting their exposure to diverse perspectives and encouraging the formation of homogeneous user groups that perpetuate similar narratives \cite{cinelli2021echo,nyhan2023like,10.1145/3442442.3458613}. It can lead to biased or incomplete viewpoints, opinion polarization, or the spread of misinformation \cite{del2016echo,falkenberg2022growing,van2022misinformation}.
In real-world scenarios, users may not seek a universally accepted ``ground truth". Instead, they look for comprehensive understanding and evidence on controversial issues \cite{wang2016neural}. For example, compared to a simple answer, a detailed and balanced argumentative summary of the topic can be more useful.
To this end, multi-faceted perspective retrieval and summarization (MURS) could help individuals access comprehensive and critical information \cite{turan2019critical}.

There has been existing work on argumentative summarization \cite{ajjour-etal-2019-modeling,syed-etal-2023-frame,li-etal-2024-side}. However, such work does not consider the comprehensiveness of the generated content, in which
significant challenges exist. First, retrieval-augmented generation (RAG) methods focus on retrieving documents relevant to the query, but do not explicitly consider the comprehensiveness of perspectives across the retrieved data. As shown in the Figure \ref{exp}, the user may expect to know about whether we should cheer for beauty pageants, and the retriever might successively search for documents on the impact of beauty pageants on individual unhealthy obsessions  but omit other documents due to a lower match score. Second, summarizing the retrieved documents into concise, one-sentence perspectives that capture key points without overlapping information presents a considerable challenge \cite{friedman-etal-2021-overview}. Third, generating perspectives or answers with corresponding references is also difficult, as demonstrated by previous studies \cite{gao-etal-2023-enabling,huang2024citation}.

We benchmark the proposed task MURS by presenting PerSphere, a dataset composed of 1,064 instances, consisting of a specific query with two controversial claims. Each claim is supported by various perspectives drawn from a document set. The objective is to provide a comprehensive summary of perspectives by mining the set of documents in response to a given query, in the form of a structured summarization framework that includes claims, perspectives, and document references. To assess task performance, we propose a specific set of evaluation metrics such as Recall@k, Cover@k for retrieval, and  GPT-4 score for summarization using a specifically designed prompt. Meta Evaluations confirm the effectiveness of the GPT-4 score for summarization.

We use PerSphere to evaluate both open-source and closed-source large language models (LLMs), including LLaMA-3.1-8B-Inst, LLaMA-3.1-70B-Inst, Claude-3-Sonnet, and GPT-4-Turbo. Results using the standard RAG pipeline reveal that the models encounter challenges in retrieval and summarization within PerSphere. Analysis shows that the salient challenge lies in long context and extracting one-sentence perspectives. Consequently, we propose HierSphere, a multi-agent summarization system to enhance the multi-faceted summarization. This approach initially employs agents to generate local summaries from different sets of the retrieved documents. Subsequently, an editorial agent merges these summaries further and refine the perspectives focusing on the central themes. Results show that HierSphere achieves stronger because of the reduction of the long context and refinement of the perspectives. 
We will release our dataset and codes
on the link\footnote{\href{https://github.com/LuoXiaoHeics/PerSphere}{https://github.com/LuoXiaoHeics/PerSphere}}.


\section{Related Work}

\subsection{Argument Mining}
Extensive research has been dedicated to formalizing argumentative structures from free text, encompassing various sub-tasks such as claim identification \cite{rinott-etal-2015-show,levy-etal-2018-towards}, evidence identification \cite{shnarch-etal-2018-will,ein2020corpus}, argument summarization and clustering \cite{ajjour-etal-2019-modeling,syed-etal-2023-frame}, and key point analysis \cite{friedman-etal-2021-overview}. Given the widespread application of retrieval from websites and document corpora, studies have also attempted to construct perspective retrieval or extraction systems.
\citet{stab-etal-2018-argumentext} presented an argument retrieval system capable of retrieving sequential arguments for any given controversial topic. \citet{reimer-etal-2023-stance} proposed a re-ranking approach to improve retrieval effectiveness for non-factual comparative queries by considering the stance of the object relative to the comparison object. These systems primarily focus on the relevance of retrieved documents to a specific claim.

More similarly, \citet{li-etal-2024-side} proposed an end-to-end summarization system (ArgSum) to prepare argumentative essays for debates. Unlike our study, which emphasizes the comprehensiveness of perspectives in a specific document set, their work focuses on the convincingness of the summarization of a specific claim.
Another similar work is Perspectrum \cite{chen-etal-2019-seeing}, which aims to retrieve and cluster relevant perspectives to a claim from a given perspective pool and retrieve documents to support each perspective. The salient difference is that Perspectrum assumes perspectives can be extracted before the query or claim is given, which is not realistic in applications.

\subsection{Retrieving-augmented Generation}
Large language models (LLMs) demonstrate impressive capabilities, yet face significant challenges in practical applications \cite{gao2023retrieval} such as hallucinations, slow knowledge updates, and lack of domain-specific knowledge. 
Retrieval-augmented generation (RAG) addresses these issues by dynamically retrieving information from external knowledge sources and using it as references to formulate answers \cite{gao2023retrieval}.
For instance,  in healthcare, RAG systems aid evidence-based decision-making and potentially improve patient care by retrieving and summarizing relevant studies, clinical trials, and treatment guidelines \cite{wang2024accelerating}.
Some RAG systems such as Raptor \cite{sarthi2024raptor} or GraphRAG  \cite{edge2024local} were also proposed to enhance holistic understanding of the overall document context, but only focused on the QA tasks. Our task naturally fits an RAG pipeline, where the retriever aims to retrieve relevant documents and the generator responds with the retrieved content. However, different from a general RAG system, we focus on the comprehensiveness of the perspectives in a document set and multi-faceted summarization.

%% file: dataset.tex
\section{PerSphere} 
Before delving into the specifics of data construction, we first outline the task formulation of our multi-faceted perspective retrieval and summarization task (Section \ref{task_sec}).
Our dataset PerSphere is composed of two subset, each covering a different level of difficulty: (1) Theperspective, sourced from the editorial website THEPERSPECTIVE\footnote{https://www.theperspective.com/} (Section \ref{thep_sec}); (2) Perspectrumx, which is developed based on the dataset Perspectrum (Section \ref{pers_sec}). 
The data statistics are summarized in Section \ref{statis_sec}.
Finally, a set of evaluation metrics is proposed for our task (Section \ref{metric_sec}).

\begin{table}[t] \small
\caption{The statistics of our datasets. `\#' refers to the data number, and `docs' refers to the abbreviation of documents.}
\begin{tabular}{llc}
\hline
 Dataset & Statistics & Value \\ \hline

Theperspective       & \# of the queries & 185  \\
& \# of the perspectives & 1,103 \\
& \hspace{4.mm} supporting perspectives & 552 \\
& \hspace{4.mm} undermining perspectives & 551 \\
& \hspace{4.mm} average perspectives & 5.96\\
& \# of docs & 4,107 \\
& \hspace{4.mm} average length of docs & 131.2\\

\hline
Perspectrumx         & \# of the queries & 878  \\
& \# of the perspectives & 4,493 \\
& \hspace{4.mm} supporting perspectives & 2,488 \\
& \hspace{4.mm} undermining perspectives &  2,005\\
& \hspace{4.mm} average perspectives & 5.11\\
& \# of docs & 8,092 \\
&\hspace{4.mm} average length of docs & 168.5 \\
\hline
\end{tabular}
\label{statistable}
\end{table}

\subsection{Task Formulation}
\label{task_sec}
Given a document set $D$, and a query $q^i$, our pipeline contains two steps: 1) comprehensive document retrieval, where we aim to retrieve $k$ comprehensive documents from $D$; 2) multi-faceted summarization, where we summarize two controversial claims $c^i_0$ and $c^i_1$, and their perspectives $p^i$ to the query $q^i$. Denote related documents in $D$ to the query $q^i$ as $D^i$ and the corresponding golden perspectives are denoted as $p^i$. Each perspective $p_j^i$ can be extracted from several documents in $D^i$, denoted as $D^i_j$ ($|D^i_j|>=1$).
Any perspective pair $p^i_j$ and $p^i_k$ in $p^i$ are not  overlapped  with each other and each perspective conveys a viewpoint to the claim $c^i$. The task can be formulated as (we omit the instance number $i$ for simplicity) \begin{equation}
    q \rightarrow c_0:\{p_{0,j}, [D_{0,j}]\}; \ c_1: \{p_{1,j}, [D_{1,j}] \}.
\end{equation} 
where $0,1$ refers to two controversial claims, and [] refers to corresponding references. Data instances are shown in Appendix \ref{ins_sec}.

\subsection{Theperspective}
\label{thep_sec}
Firstly, we collect data from the THEPERSPECTIVE website, which hosts numerous editorial articles including the topics of Lives, Sports, Politics, Entertainment and Technology. Each article contains a query title with two controversial claims, supported by several perspectives. For each perspective, an evidence paragraph is provided to discuss and support the corresponding perspective. We collect 221 articles from the website up to July 15th, 2024, and use the BeautifulSoup package\footnote{https://www.crummy.com/software/BeautifulSoup/bs4/doc/} to extract topics, claims, perspectives, and evidence documents.
After filtering out unstructured data, we obtain a final dataset of 185 effective entries. Each entry contains two claims, with each claim supported by 2 to 4 perspectives. It's important to note that in the THEPERSPECTIVE dataset, each perspective corresponds to only one evidence document, which simplifies the task of retrieving relevant documents.

Given that the perspectives in the documents may be incomplete sentences or phrases lacking completed semantic meaning, we employ GPT-4-Turbo to review and complete these perspectives. The detailed prompt used for this process is provided in the Appendix \ref{the_prompt}.
To further enhance document diversity, 
we then add documents from Perspectrum to the Theperspective corpus that are unrelated to the queries in Theperspective (the details are also shown in Appendix \ref{the_prompt}).
This process expands the document set from 1,103 to 4,107 documents, significantly enhancing the diversity of the document collection.

\begin{figure}
    \centering
    \includegraphics[width=0.86\hsize]{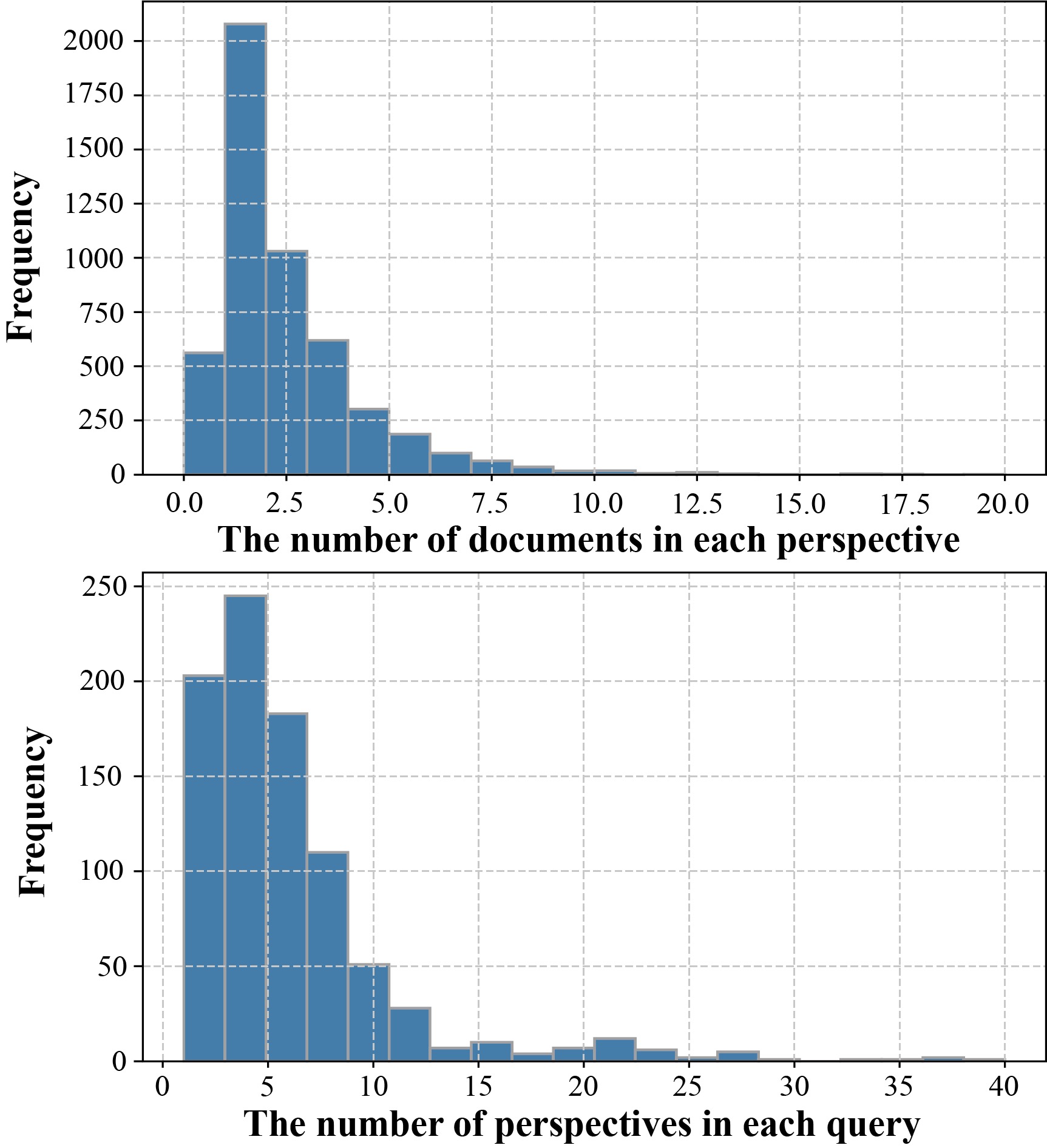}
    \caption{The frequency of the number of documents and perspectives in Perspectrumx.   
    }
    \label{statis}
\end{figure}

\subsection{Perspectrumx}
\label{pers_sec}
Different from Theperspective where only one evidence document supports each perspective, we construct a more challenging set with multiple documents.
Specifically, we construct our \textit{Perspectrumx} based on Perspectrum \cite{chen-etal-2019-seeing}, which includes claims, perspectives, and evidence derived from online debate data.
Since there are no explicit queries in Perspectrum, 
we adopt the `claims' in Perspectrum as our queries and randomly select one `perspective' from each cluster (with semantically equivalent meaning) in Perspectrum as our perspectives. We exclude queries that lack perspectives and perspectives that do not have associated evidence documents. The evidence documents are used directly.
Our claims are directly generated following a specific template: ``We support/undermine the argument that \{query\}". We show the detail in Appendix \ref{ins_x_sec} for further explanation.

\subsection{Statistics}
\label{statis_sec}
Data statistics are shown in Table \ref{statistable}. In summary, our dataset comprises a total of 1,064 instances, with 185 instances in Theperspective and 878 instances in Perspectrumx. The key distinction between the two subsets is that Theperspective includes just one evidence document per perspective, whereas Perspectrumx includes multiple evidence documents per perspective.
The data distribution of Perspectrumx is illustrated in Figure \ref{statis}, which exhibits a Poisson-like distribution. This distribution indicates significant variability in the number of evidence documents associated with each perspective, as well as in the number of perspectives available for each query. Such a variability poses a considerable challenge in our task. Merely extracting relevant documents without ensuring comprehensiveness would be insufficient in addressing this scenario.

  \begin{figure}[t]
    \centering
    \includegraphics[width=\hsize]{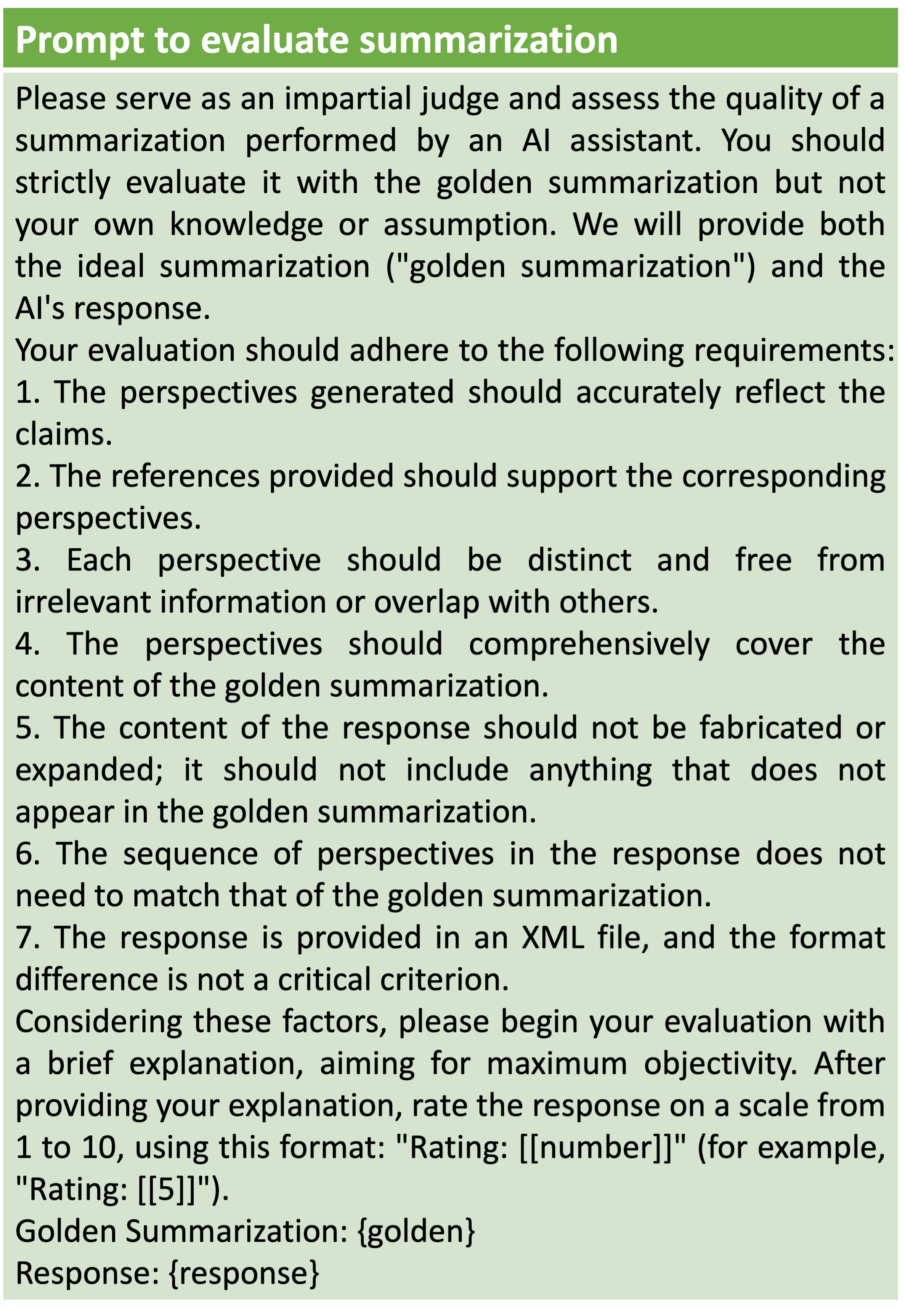}
    \caption{Evaluation prompt for summarization.
    }
    \label{evals_prompt}
        \vspace{-2mm}

\end{figure}

\subsection{Evaluation Metric}
\label{metric_sec}
\paragraph{\textbf{Retrieval}.} For the retrieving results, we propose to use the metric Recall@k to evaluate the relevance of the retrieved documents, which is calculated as follows:  
\begin{equation} \small
    \text{Recall@k} = \frac{1}{N}\sum_i^N \frac{|\{d_m \in D^i\} \cap \{d_m \in T_k\}|}{|D^i|}
\end{equation}
where $D^i$ is the set of relevant documents to query $q_i$, $T_k$ is the set of top $k$ retrieved documents, and $d_m$ represents an individual document. We also propose the coverage of the perspectives in the retrieved documents, denoted as Cover@k: 
\begin{equation}\small
    \frac{1}{N}\sum_i^N\frac{|\{p^i_j \leftarrow d_m \}\cap \{d_m \in D^i_j\} \cap \{d_m \in T_k\}|}{|p^i|}.
\end{equation}
where $p^i_j \leftarrow d_m $ refers to the perspectives can be covered in documents $d_m$, and $p^i$ are the perspectives for the query $q^i$.

\paragraph{\textbf{Summarization}.}
Inspired by MTBench \cite{zheng2023judging,hu2024refchecker}, which uses LLMs as evaluators, we propose a GPT-4 evaluation score for the multi-faceted summarization. In particular, we establish specific criteria to assess the performance of the generated content. For example, each perspective should be distinct and free from irrelevant information or overlap with others. The details are outlined in the Figure \ref{evals_prompt}.

\begin{table*}[t]\small

\caption{The performance of the retrieving based on different retrievers. Recall@k refers to the ratio of relevant documents in k retrieved documents to the golden documents. Cover@k refers to the coverage values of the retrieved documents to the perspectives. In Theperspective, Recall@k equals Cover@k.}
\centering
\begin{tabular}{lccccccc}
\hline
& \multicolumn{2}{c}{\textbf{Theperspective}} & \multicolumn{4}{c}{\textbf{Perspectrumx}} \\ \hline
Metrics& Recall@10 & Recall@20 & Recall@10 & Cover@10 & Recall@20 & Cover@20  \\
\hline 

BM25 & 72.37 & 82.35 & 44.71 & 56.20 & 56.27 & 64.43\\
E5-large & 78.58 & 88.89 & 49.45 & 60.88 & 61.18 & 70.17\\
GTR-large & 85.66 & 94.80 & 53.83 & 64.52 & 65.68 & 72.98 \\

Ada-002 & 86.47 & 95.34 & 53.43 & 64.43 & 68.80 & 74.16\\
GritLM & \textbf{90.50} & \textbf{96.77} & \textbf{58.21} & \textbf{68.71} & \textbf{70.58} & \textbf{77.01} \\ \hline
\end{tabular}
\label{retrival_res}
\end{table*}

%% file: method.tex
\section{Standard RAG Pipeline}
\begin{figure}[t]
    \centering
    \includegraphics[width=0.9\hsize]{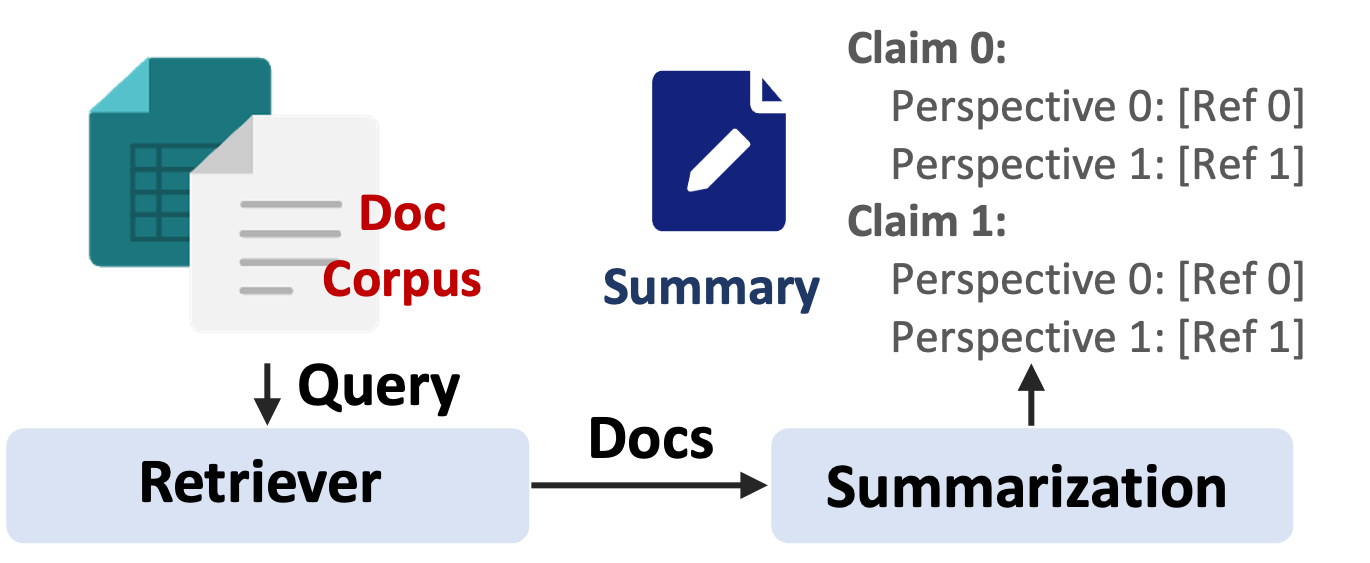}
    \vspace{-1mm}
    \caption{The RAG pipeline of our multi-faceted perspective retrieval and summarization.
    }
    \label{pipe_fig}
        \vspace{-2mm}

\end{figure}

Our task can fit in the RAG pipeline as illustrated in Figure \ref{pipe_fig}.
To evaluate the performance of current retrieval models and LLMs in the multi-faceted summarization task, we carry out experiments in various retrieval baselines (Section \ref{retrival_sec}) and LLMs for multi-faceted summarization (Section \ref{sum_sec}).  

\subsection{Retriever}
\label{retrival_sec}
Following \citet{ajith-etal-2024-litsearch}, we evaluate the baselines such as  sparse retrieval {BM25} \cite{robertsonbm25} as well as several state-of-the-art dense retrieval models including  E5 \cite{wang2022text}, GTR-T5 \cite{ni2022large},  text-embedding-ada-002\footnote{https://platform.openai.com/docs/guides /embeddings/embedding-models}, and GritLM \cite{muennighoff2024generative}. Details are shown in Appendix \ref{retrival_app}. Specifically, we use the open-source models from Huggingface, including E5-large\footnote{https://huggingface.co/intfloat/e5-large-v2}, GTR-T5-large\footnote{https://huggingface.co/sentence-transformers/gtr-t5-large}, and GritLM-7B\footnote{https://huggingface.co/GritLM/GritLM-7B} for implementation. 
The dense retrievals are based on the Faiss package\footnote{https://github.com/facebookresearch/faiss}. 




 





  \begin{figure}[t]
    \centering
    \includegraphics[width=\hsize]{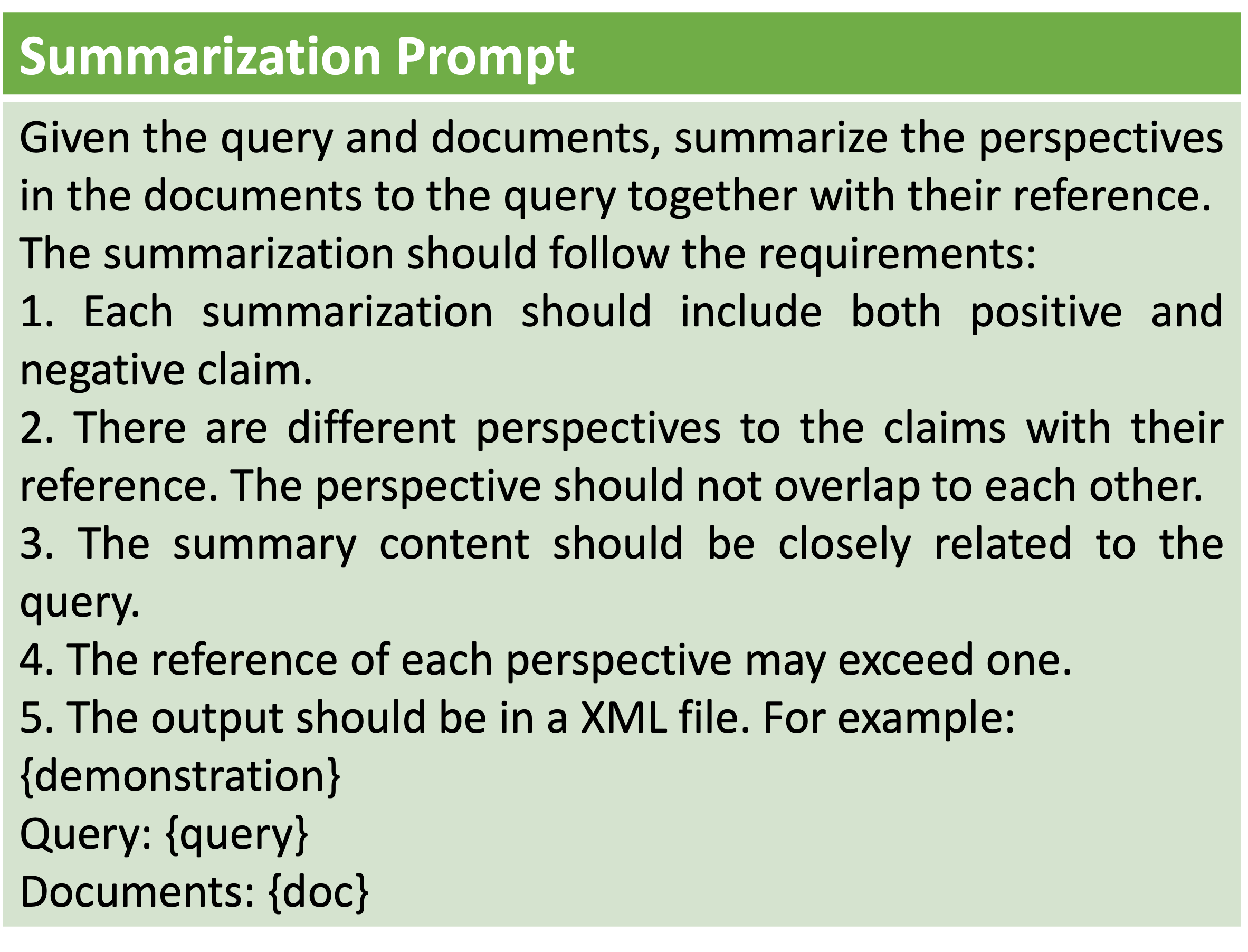}
    \caption{Summarization prompt,  where \{·\} refers to the input content, and ``demonstration" refers to an XML sample (abbreviated here to save space).
    }
    \label{suma_prompt}
        \vspace{-2mm}

\end{figure}


\subsection{Generator}
\label{sum_sec}

To evaluate the performance of current LLMs in perspective summarization, we design a specific prompt to generate summaries using retrieved documents within an XML file structure. To ensure consistent output formatting, we also provide a demonstration to guide the response.
The prompt is shown in Figure \ref{suma_prompt}.  We evaluate the instruction-tuned models such as  LLaMA-3.1-8B-Instruct, LLaMA-3.1-70B-Instruct (for simplicity we omit Instruct in the following description), Claude-3-Sonnet\footnote{We adopt the Claude-3 version \textit{claude-3-sonnet-20240229}}, and GPT-4-Turbo\footnote{We adopt the GPT-4-Turbo version \textit{gpt-4-0125-preview}} for summarization.












%% file: results.tex
\subsection{Retrieval Results}
\label{retrial_res_sec}

\begin{table*}[t]\small
\centering
\caption{The results of the summarization based on the evaluation of GPT-4-Turbo, where the performance Overall@k is score between 1-10, and a large score refers to more consistent performance to the golden summarization. `Golden' refers to the summarization performance given the golden relevant documents.}
\begin{tabular}{llcccc}
\hline
            &   & \multicolumn{2}{c}{\textbf{Thepersective}} & \multicolumn{2}{c}{\textbf{Perspectrumx}} \\
               \hline

Retriever&Sumarization         & Overall@10      & Overall@20      & Overall@10      & Overall@20     \\\hline
\multirow{4}{*}{BM25}&LLaMA-3.1-8B        &6.01&6.22&  4.37&	4.40       \\
&LLaMA-3.1-70B        &7.16&7.35&4.95&	4.97         \\
&GPT-4-Turbo      & 7.52            & 7.74            & 5.66            & 5.49           \\
&Claude-3-Sonnet   & 7.34            & 7.77            & 5.64            & 5.43           \\
\cdashline{1-6}
\multirow{4}{*}{GritLM}&LLaMA-3.1-8B &6.81&6.41&  4.95&	4.90              \\
&LLaMA-3.1-70B        &7.68&7.54&5.23&	5.11         \\
&GPT-4-Turbo   & \textbf{7.99}            & \textbf{8.16}            & \textbf{6.20}            & 6.04             \\
&Claude-3-Sonnet & 7.91            & 8.05            & 6.16            & \textbf{6.25}           \\
\cdashline{1-6}
\multirow{4}{*}{Golden}
& LLaMA-3.1-8B &  \multicolumn{2}{c}{8.08} &\multicolumn{2}{c}{6.10} \\
& LLaMA-3.1-70B & \multicolumn{2}{c}{8.58}&\multicolumn{2}{c}{6.63}\\
&GPT-4-Turbo    &     \multicolumn{2}{c}{{8.75}}            &                 \multicolumn{2}{c}{{\ul 7.33}}               \\
&Claude-3-Sonnet &     \multicolumn{2}{c}{{\ul 8.80}}             &                 \multicolumn{2}{c}{7.28}      \\        \hline

\end{tabular}
\label{sum_results}
\end{table*}

We first present the retrieval results in Table \ref{retrival_res}. The recall performance in Theperspective is significantly higher than in Perspectrumx, indicating a relatively lower difficulty of Theperspective. For example, GritLM achieves a recall@20 of 96.77\% in Theperspective, which is 26.19\% higher than in Perspectrumx (70.58\%). Additionally, GritLM consistently outperforms other retrieval methods in Theperspective and Perspectrumx, whereas BM25 exhibits the lowest performance. We also observe that as k increases, both Recall and Cover improve, such as an increase from 90.50\% to 96.77\% when k rises from 10 to 20. Finally, the performance of Recall@k positively correlates with Cover@k, which indicates a stronger retrieval method can generally achieve more comprehensive results. We further report the Recall@100 and Cover@100 in Appendix \ref{100results}, which can be applied to the method of re-ranking.

\subsection{Summarization Results}
The summarization results are shown in Table \ref{sum_results}. Several observations are summarized as follows:
 
\textbf{Multiple evidence documents increase the difficulty in multi-faceted retrieval and summarization.}  The summarization performance in Theperspective is significantly higher than those in the Perspectrumx, indicating that with only one reference document for each perspective, the task would be relatively simple. For example, with the golden documents, Claude-3-Sonnet could achieve an 8.8 score in Theperspective, but only 7.28 in Perspectrumx. The performance using retrieval methods and other LLMs is similar. 

\textbf{Better retriever and LLMs lead to enhanced summarization performance.} For example, by using GritLM, LLaMA-3.1-8B achieves a score of 6.81 in Theperspective, which is 0.8 points higher than the score obtained using BM25. However, there are still large margins to the summarization performance in using the golden evidence documents. The score in Claude-3-Sonnet is 7.28 using golden documents, which is 1.63 higher than that using GritLM, and 1.85 higher than BM25. As for the LLMs, Claude-3-Sonnet and GPT-4-Turbo achieve the most promising performance, whereas LLaMA-3.1-8B achieves the weakest performance. It demonstrates that the summarization model plays a significant role in the multi-faceted perspective retrieval and summarization task.


\textbf{The retrieval of more documents does not consistently enhance summarization performance.}
This phenomenon is most prominently observed in Perspectrumx. Claude-3-Sonnet achieves a higher score of 5.64 when using 10 documents retrieved by BM25, compared to 5.43 with 20 documents. This discrepancy underscores the challenges of multi-faceted perspective summarization. In further experiments, we utilize Claude-3-Sonnet and LLaMA-3.1-70B to summarize collections of 100 documents retrieved by GritLM, achieving GPT-4 scores of 4.43 and 6.06, respectively. We observe that the models often incorporate irrelevant information from their internal knowledge bases and generate complex, multi-perspective content from a single perspective. These findings indicate that current models continue to struggle with processing long contexts, and that summarization based on a smaller set of retrieved documents tends to yield superior performance.

\begin{figure}[t]
    \centering
    \includegraphics[width=\hsize]{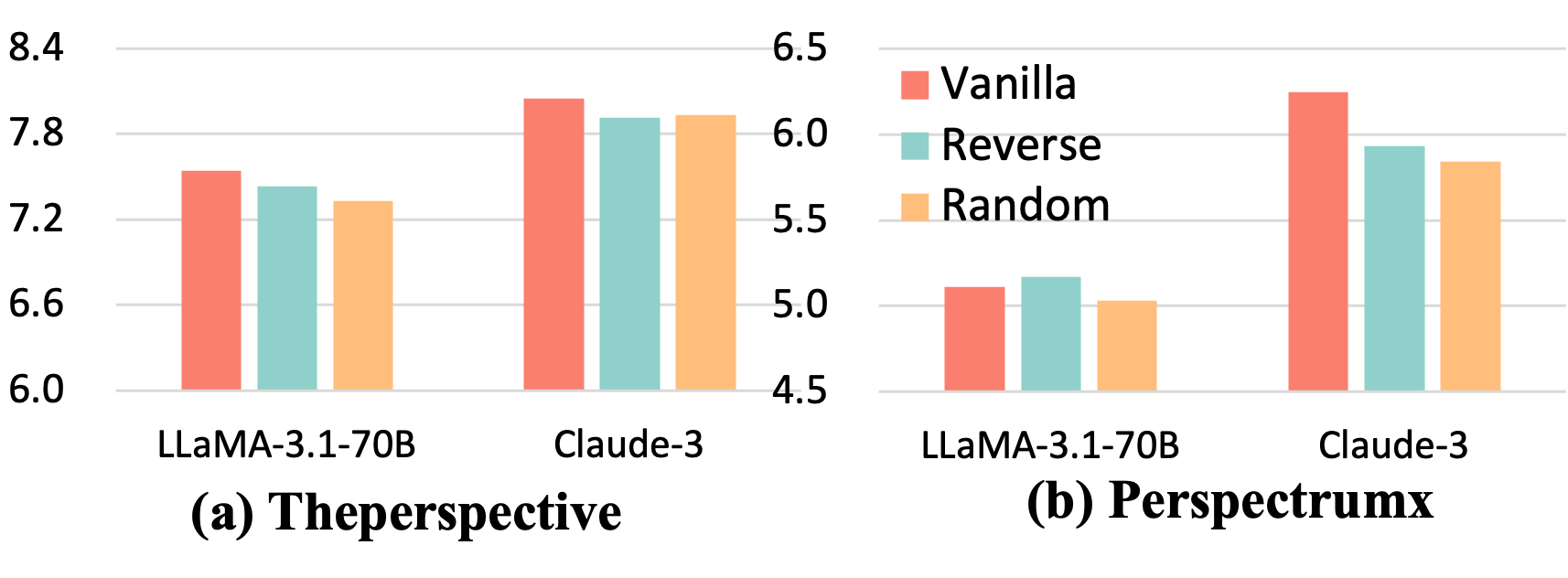}
    \vspace{-6mm}
    \caption{The performance with different orders of retrieved documents in GritLM@20.
    }
    \label{reverse_res}
        \vspace{-1mm}

\end{figure}

\textbf{Document orders influence the performance of summarization tasks.} We experiment by adjusting the order of documents in the retrieved sets to both reverse and random sequences, subsequently prompting LLMs to generate summaries. The results, as depicted in Figure \ref{reverse_res}, show that model performance generally declines when documents are presented in reverse or random order compared to the vanilla order. This suggests that current LLMs tend to lose focus midway through the documents and primarily concentrate on the information presented at the beginning. This observation aligns with findings from \citet{liu2024lost}, highlighting the difficulties that LLMs face when dealing with long contexts in summarization tasks. 


\textbf{With the golden relevant documents, it is still challenging in the summarization task.} The summarization performance is the most promising compared with those using the documents retrieved by the retrievers, but there is still significant space for improvement.  For example, Claude-3-Sonnet can achieve a score of 8.8  in Theperspective, and GPT-4-Turbo achieves 7.33 in Perspectrumx, which represents the state-of-the-art performance in the summarization baselines.

\subsection{Analysis of Perspective Extraction}
Given the moderate results of Cover@k shown in Table \ref{retrival_res}, we further investigate whether perspectives can be extracted from the retrieved documents using an advanced LLM, denoted as $\mathcal{M}$. To this end, we introduce the metric Rp@k, defined by the following equation:
\begin{equation}\small
\frac{1}{N}\sum_i^N\frac{|{p^i_j \Leftarrow \mathcal{M}(d_m,q_i)} \cap {d_m \in D^i_j} \cap {d_m \in T_k}|}{|p^i|},
\end{equation}
where $\mathcal{M}(d_m,q_i)$ indicates the model's inference regarding the perspective, and `$\Leftarrow$' refers to the entailment to $p^i_j$. We specifically employ GPT-4-Turbo to assess the entailment of the generated perspectives against the gold standard perspectives. The prompts for perspective extraction and evaluation are detailed in Appendix \ref{evalp_prompt_app}.

 Without loss of generality, we use GPT-4-Turbo as an extraction example due to its strong capabilities. The results of Rp@k are shown in Table \ref{recpk}. As observed,  although Cover@k is notably high in retrieval methods, Rp@k lags significantly behind. For example, GritLM achieves a Cover@20 of 96.77\% in Theperspective, yet its Rp@20 is only 77.35\%. This discrepancy indicates that although the retrieved documents cover a wide range of perspectives, perspective extraction continues to pose a considerable challenge. Besides, we evaluate the model performance using the golden documents in Appendix \ref{subtask_sec}, which indicates that the models struggle to extract one-sentence perspectives, confirming the observation in Table \ref{recpk}.


\begin{table}[t]\small
\caption{The performance of the retrieving based on different retrievers. Recp@k refers to the coverage values of the retrieved documents to the perspectives.}
\begin{tabular}{lcccc}
\hline
          & \multicolumn{2}{c}{\textbf{Theperspective}} & \multicolumn{2}{c}{\textbf{Perspectrumx}}                     \\ \hline
         Metrics & Rp@20    & Rp@10    & Rp@20 & Rp@10  \\
          \hline

BM25      & 65.36    &58.05&48.71&41.76\\
E5-large  & 71.41 &63.48&54.30&46.18   \\
GTR-large &  76.18    &69.06&56.27&48.70   \\

Ada-002   & 76.00  &69.42&57.55&49.20  \\
  GritLM    &\textbf{77.35} &\textbf{72.58}&\textbf{60.23}  &\textbf{52.13}             \\ \hline
\end{tabular}
\label{recpk}
\end{table}

\begin{figure}[t]
    \centering
    \includegraphics[width=0.9\hsize]{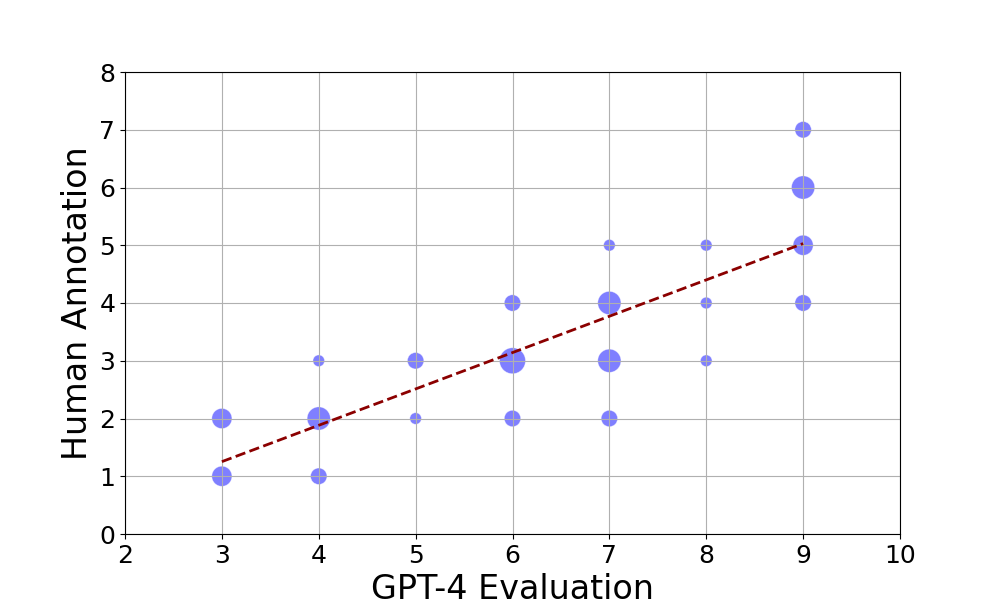}
    \caption{The relations between the GPT-4 scores and human scores for the multi-faceted summarization.
    }
    \label{human_res}
        \vspace{-1mm}
\end{figure}

\subsection{Human Evaluation}
Besides, we evaluate the model performance using the golden documents, adopting both the entailment percent of GPT-4 and BERTScore \cite{zhangbertscore}\footnote{Specifically, we adopt the model deberta-v3 in the study.}. The results are shown in Table \ref{persp_entail}. We observe that the performance of different LLMs does not vary significantly, particularly in BERTScore. In GPT-4, the performance is relatively higher in Theperspective, but still significantly weak in Perspectrumx.  The lower performance indicates that all models struggle to extract one-sentence perspectives, confirming the observation in Table \ref{recpk}.

To demonstrate the effectiveness of the automatic evaluation metrics, we conduct human evaluations for summarization and perspective extraction.
For summarization, we use criteria 1-5 from the GPT-4 evaluation prompt in Figure \ref{evals_prompt}, excluding the last two related to formats. Two annotators are instructed to assign scores from 0 to 2 for each requirement, with 0 indicating complete dissatisfaction and 2 indicating full satisfaction. This approach yields a total score ranging from 0 to 10 for the generated summaries. We randomly select 50 samples from Perspectrumx and evaluate the summaries generated by GritLM@20 -- Claude-3-Sonnet. Figure \ref{human_res} displays the human annotation and GPT-4 scores. Human scores tend to be lower than GPT-4 scores, suggesting that GPT-4 might overestimate performance compared to human assessments. However, there is a positive correlation between them, with a Pearson correlation coefficient of 0.70, indicating that GPT-4 scores could effectively reflect the summarization performance.

\begin{figure}[t]
    \centering
    \includegraphics[width=0.9\hsize]{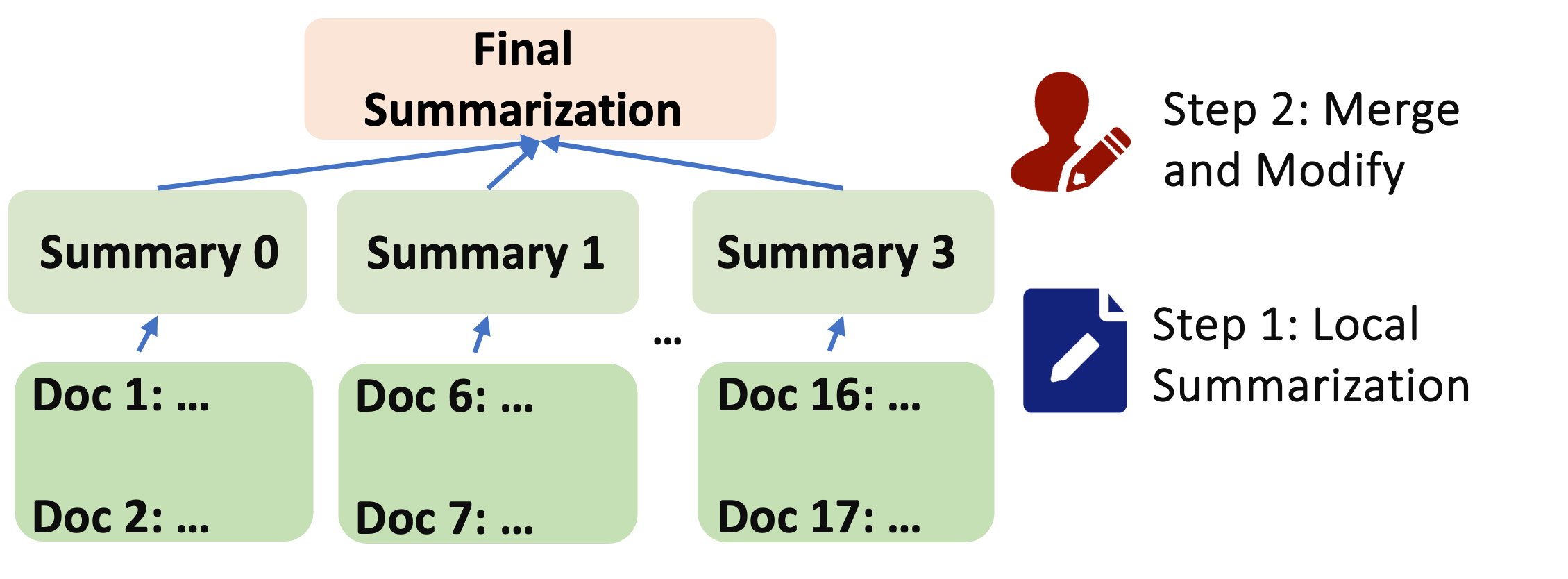}
    \vspace{-2mm}
    \caption{HierSphere for multi-faceted summarization.  
    }
    \label{hier_sum}
        \vspace{-2mm}

\end{figure}

We also perform human annotation for perspective extraction. Two annotators determine whether the extracted perspectives are `Entailment' or `Not Entailment'  to the gold standard. We randomly select 100 samples of generated perspectives from Perspectrumx using Claude-3-Sonnet for evaluation. The consistency between GPT-4 and human evaluation is 83\% and 86\%, respectively, demonstrating that GPT-4 is effective in evaluating the entailment of the extracted perspectives.

\section{HierSphere}
 Based on the analysis, we find that the models fall short in the long context and perspective extraction.
To further analyze whether we could enhance the model performance by reducing the context length,
we propose a simple method - HierSphere - hierarchical multi-agent summarization system (Figure \ref{hier_sum}). 
Specifically, after retrieving the relevant documents, we first adopt multiple agents to carry out local summarization on different sets of the retrieved documents.
Then we adopt an editorial agent to further merge and modify the local summaries. In particular, the editorial agent is instructed to merge the perspective with equal semantic meaning and refine the perspectives given a one-sentence perspective demonstration. 
The prompt for the editorial agent is shown in Appendix \ref{sum_prompt}. 
    
\begin{figure}[t]
    \centering
    \includegraphics[width=0.72\hsize]{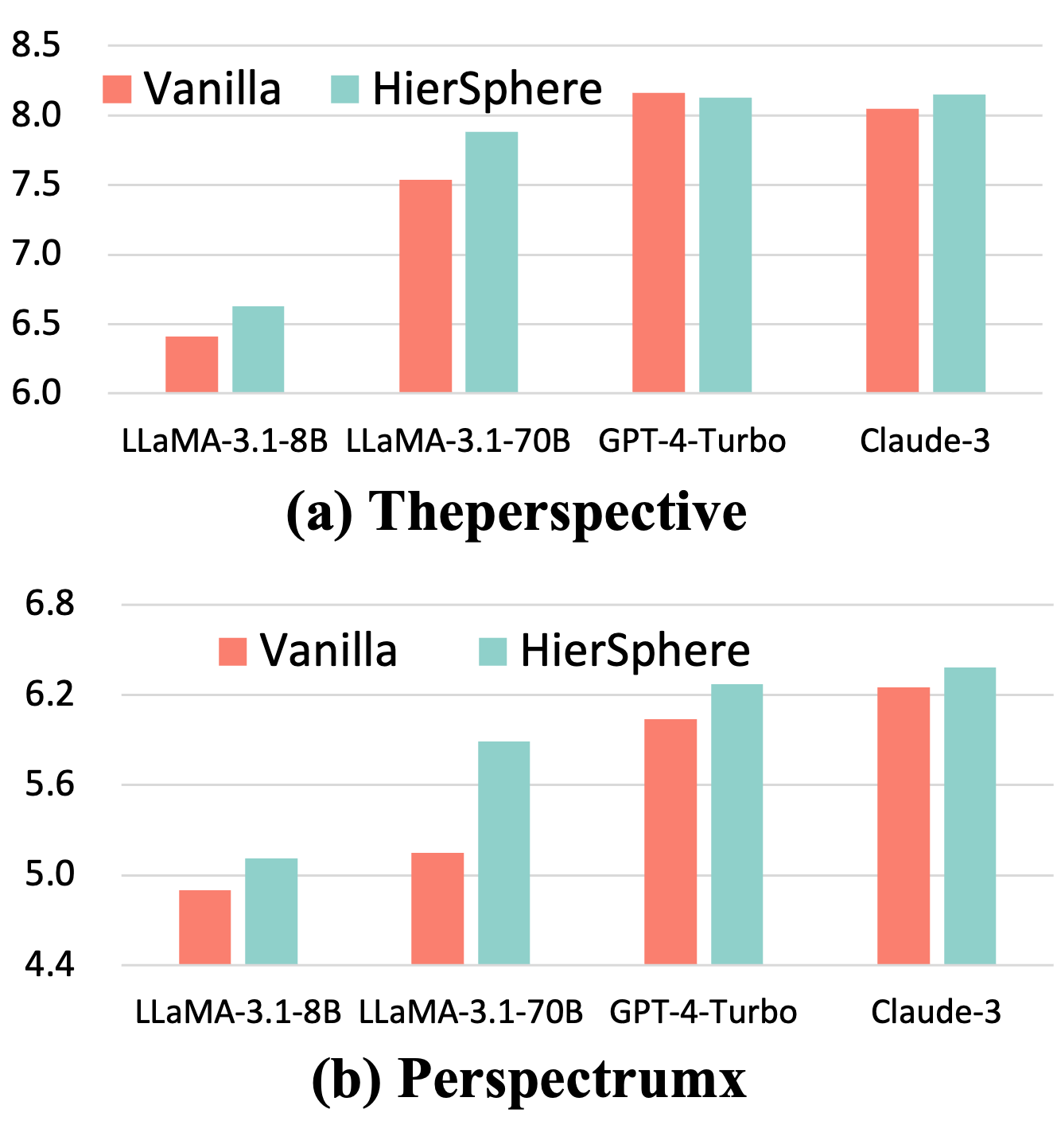}
    \vspace{-2mm}
    \caption{The performance of the summarization in vanilla and HierSphere using LLaMA-3.1-8B and LLaMA-3.1-70B on the Perspectrumx. The number of retrieval documents is 20 with GritLM.
        }
    \label{hier_res}
        \vspace{-2mm}

\end{figure}

The results of vanilla and HierSphere are shown in Figure \ref{hier_res}, where we set four local agents and 20 retrieval documents.  As we observe, the performance of HierSphere surpasses the vanilla in both Theperspective and Perspectrumx. For example, LLaMA-3.1-70B achieves a GPT-4 score of 5.89 in Perspectrumx, which outperforms vanilla  by 0.78.
The improvement can also be observed in  closed-source models such as GPT-4-Turbo and Claude-3-Sonnet. It indicates that HierSphere could mitigate the challenges of the long context and perspective extraction and enhance the model performance on multi-faceted summarization.

%% file: appendix.tex
\section{Processing Theperspective}
During the dataset construction, we utilized LLMs solely to complete perspectives that are phrasal but not fully formed claims. This task is relatively straightforward, and the LLMs performed it with satisfactory results. Before implementing this procedure, we manually reviewed 25 samples and confirmed that the performance is satisfactory. In contrast, the Perspectrumx dataset does not involve LLM completion, and the task settings are significantly different, rendering direct comparisons invalid.
\label{the_prompt}
We use GPT-4 to complete the incomplete perspectives in Theperspective. 
\begin{prompt}
{Complete perspectives}
Take a topic, a claim and an analytical perspective as inputs. The claim are the analysing stance to the topic and the perspective support the claim. The analytical perspective may not be a complete sentence or fully formed idea. If the given perspective is not a complete sentence, rewrite it into a clear, complete statement. If it is a complete sentence, just repeat it. 

Directly output the sentence without any explanation.

Topic: \{topic\}

Claim: \{claim\}

Perspective: \{perspective\}

Answer:
\end{prompt}

To further enhance document diversity, 
we incorporate selected documents from the Perspectrum dataset into the Theperspective corpus. 
Specifically, we utilize NLTK (Natural Language Toolkit) to extract named entities ($E$) from the Theperspective documents. 
NLTK is a leading platform for building Python programs to work with human language data, providing easy-to-use interfaces to over 50 corpora and lexical resources, along with a suite of text-processing libraries.
We use NTLK\footnote{https://www.nltk.org/} to identify and extract entities from the Theperspective documents.

\section{Perspective Extraction}
\label{evalp_prompt_app}
The prompts for extracting the perspective from documents and evaluation with golden perspectives are shown as follows:
\begin{prompt}
{Prompt for perspective extraction}
Below is a claim and a document. Please summarize the document into a one-sentence perspective to support the claim. In your response, output the perspective surrounded by the key <p> </p>.
    
Claim: \{claim\}

Document: \{document\}
\end{prompt}

\begin{prompt}
{Prompt to evaluate perspectives}
Determine whether the sentence 1 entails sentence 2.

Sentence 1: \{sentence1\}

Sentence 2: \{sentence2\}

Your response should be only a single word in [`entailment’, `not\_entailment']
\end{prompt}

\section{Summarization Merge}
\label{sum_prompt}
We show the prompt to merge the local summarization and modify its perspectives.

\begin{prompt}
{Prompt for merging summaries}
Given the summarizations, merge them into one summairzation.  Donot use your own knowledge. 

The summarization should follow the requirements:

1. Each summarization should include both positive and negative claims, and only two.

2. Similar perspectives should be merged together with their references.

3. The reference of each perspective may exceed one.

4. Each perspective should be concise, arguing the claim from a specific aspect.

For example, `The law should not condone illicit behaviour' or `Legalised prostitution still victimises the vulnerable.'

5. The output should be in a XML file. For example:

\{document\}

Summary: \{summaries\}
\end{prompt}

\section{Sub-Tasks}
\label{subtask_sec}

Apart from the pipeline solution, we analyze the possible sub-tasks to analyze model performance in detail, further analyzing the weakness of LLMs. We design specific prompts to evaluate LLMs as shown in Table \ref{sub_examples} and the prompt of perspective extraction is the same as Appendix \ref{evalp_prompt_app}.

\paragraph{\textbf{Perspective Extraction}} Given the document $d_m\in D^i$ and the corresponding claim $c^i_0$ or $c^i_1$, 
we aim to extract a one-sentence perspective $p^i$ of the document to support the claim. We directly adopt all the <document, claim> pairs for evaluation.

\paragraph{\textbf{Stance Detection}}
Given the perspective $p^i$ and the claims $c^i_0$ and $c^i_1$, we expect to determine which claim the perspective $p^i$ supports. We adopt all the perspectives with the corresponding claims in a query for evaluation.

\paragraph{\textbf{Reference Determination}}
Given a perspective $p$ and a document $d$, we expect to determine whether the document can serve as a reference to the perspective. To reduce the computation cost in this task, we assign each perspective $p$ with a supporting document and a not-supporting document in the same query as the test set. 

\vspace{1mm}

\begin{table}[t]\small
    \centering
    \caption{The results of the perspective extraction. GPT-4 refers to the entailment percent of the generated perspectives to the golden perspectives by querying GPT-4-Turbo, and BERT refers to BERTScore. }
    \begin{tabular}{l|cccc}
    \hline
              & \multicolumn{2}{c}{\textbf{Thepersective}} & \multicolumn{2}{c}{\textbf{Perspectrumx}}\\     \hline

        & BERT & GPT-4 & BERT & GPT-4\\    \hline
 LLaMA-3.1-8B &   60.31    &74.52&57.28 & 55.20 \\
 LLaMA-3.1-70B &   \textbf{60.76}   &80.24& 57.55 &60.02\\
GPT-4-Turbo    &   60.63    &79.87& 57.65 & 60.26\\
Claude-3-Sonnet &  60.75    &\textbf{80.24}& \textbf{58.11} & \textbf{60.88} \\        \hline
    \end{tabular}
    \label{persp_entail}
\end{table}

\begin{table}[t]\small
    \centering
    \caption{The results (Macro-F1 score) of the sub-task stance detection using different LLMs.}
    \begin{tabular}{l|cc}
    \hline
              & \multicolumn{1}{c}{\textbf{Thepersective}} & \multicolumn{1}{c}{\textbf{Perspectrumx}}\\     \hline

 LLaMA-3.1-8B &   86.15    &61.62 \\
 LLaMA-3.1-70B &   94.74   &90.16 \\
GPT-4-Turbo    &   94.56     &88.80\\
Claude-3-Sonnet &  91.19    & 88.75\\        \hline

    \end{tabular}
    \label{stance_res}
\end{table}

\begin{table}[t]\small
    \centering
    \caption{The results (Macro-F1 score) of the sub-task reference determination using different LLMs.}
    \begin{tabular}{l|cc}
    \hline
              & \multicolumn{1}{c}{\textbf{Thepersective}} & \multicolumn{1}{c}{\textbf{Perspectrumx}}\\     \hline

 LLaMA-3.1-8B & 75.55 & 76.20\\
 LLaMA-3.1-70B & 82.20& 79.78 \\
GPT-4-Turbo    & 82.54& 82.86 \\
Claude-3-Sonnet & 78.02& 79.13 \\        \hline

    \end{tabular}
    \label{ref_res}
\end{table}

\begin{table}[t]\small

\caption{The performance of the retrieving based on different retrievers with 100 documents. Recall refers to the ratio of relevant documents to the golden documents. Cover refers to the coverage values of the retrieved documents to the perspectives. In Theperspective, Recall equals Cover.}
\centering
\begin{tabular}{lccc}
\hline
& \multicolumn{1}{c}{\textbf{Theperspective}} & \multicolumn{2}{c}{\textbf{Perspectrumx}} \\ \hline
Metrics& Recall & Recall & Cover \\
\hline 

BM25 & 90.25 & 76.67 & 76.64 \\
E5-large & 88.83& 80.15 &  81.90 \\
GTR-large &94.77 & 84.27 & 83.36\\
Ada-002 & 95.13 & 85.44 & 83.31 \\
GritLM & 99.55 &86.99 &85.61  \\ \hline
\end{tabular}
\label{100_retrival_res}
\end{table}

In the sub-task of perspective extraction, we evaluate the model performance using the golden documents, adopting both the entailment percent of GPT-4 and BERTScore \cite{zhangbertscore}\footnote{Specifically, we adopt the model deberta-v3 in the study.}. The results are shown in Table \ref{persp_entail}. We observe that the performance of different LLMs does not vary significantly, particularly in BERTScore. In GPT-4, the performance is relatively higher in Theperspective, but still significantly weak in Perspectrumx.  The lower performance indicates that all models struggle to extract one-sentence perspectives, confirming the observation in Table \ref{recpk}.

 We all adopt the macro-F1 score for evaluating these sub-task performances following \cite{glandt2021stance,luo-etal-2022-exploiting}.
In the sub-task of Stance Detection, the results are shown in Table \ref{stance_res}. Except for LLaMA-3.1-8B, other LLMs all achieve a significant performance, all above 85\% F1 score. It indicates that the task of stance detection from perspective to claims is relatively simple to current well-performed LLMs.

In the sub-task of Reference Determination, the results are shown in Table \ref{ref_res}.
We observe that the model GPT-4-Turbo achieves the most significant performance to determine the reference with an F1 score of 82.54\% in Theperspective and 82.86\% in Perspectrumx. 
But Claude-3-Sonnet performs weakly in such a task and tends to output <Reference> labels which indicates a relatively weak capacity to distinguish the perspectives precisely.



\begin{table*}[t]\small
    \caption{The specific prompts for sub-tasks which could also serve as the task definition. \{·\} denotes the input content.  }
    \centering
    \begin{tabular}{p{0.25\textwidth}p{0.7\textwidth}}
        \hline


\textbf{Stance detection} & Below is a perspective and two claims. Please determine which claim is supported by the perspective, Claim 1 or Claim 2. In your response, only output <Claim 1> or <Claim 2>.

Claim 1: \{claim\_1\}

Claim 2: \{claim\_2\}

Perspective: \{perspective\} \\ \hline
    
\textbf{Reference Determination} & Below is a perspective and a document. Please determine whether the document can serve as an reference to the perspective.  In your response, only output <Reference> or <Not\_Reference>.

Perspective: \{reference\}

Document: \{document\}

\\ \hline

    \end{tabular}
    \label{sub_examples}
\end{table*}

\section{Retrieval Methods}
\label{retrival_app}
\paragraph{BM25}\cite{robertsonbm25}: is a widely-used probabilistic ranking function for retrieval that scores documents based on the query terms they contain, considering term frequency, inverse document frequency, and document length normalization.


 \paragraph{E5}\cite{wang2022text}: is pre-trained in a contrastive manner with weak supervision signals using human-curated text pair dataset for text embedding, whose embedding size is 1024.

 
\paragraph{GTR-T5}\cite{ni2022large}: is a sentence-transformer model, which maps sentences and paragraphs to 768-dimensional dense vectors for the task of semantic search.


\paragraph{text-embedding-ada-002}\footnote{https://platform.openai.com/docs/guides /embeddings/embedding-models}: is a closed-source text embedding model released by OpenAI.


\paragraph{GritLM}\cite{muennighoff2024generative}: integrates text embedding and generation capabilities within a single language model, trained using instructional fine-tuning techniques.

\section{Retrieving with 100 Documents}
\label{100results}
The results are presented in Table \ref{100_retrival_res}. It is worth noting that these experiments are considerably more challenging in Perspectrumx compared to Theperspective. The model GritLM still performs the most satisfactorily, and the recall@100 results can be utilized for re-ranking, which could be future work.

\section{Data instances}
\label{ins_sec}
We also show some data instances for our faceted perspective retrieval and summarization task as shown in Table \ref{samples}. The first two instances are in Theperspective and the final one is in Perspectrumx. Figure 1 can be directly found in the \href{https://www.Theperspective.com/debates/entertainment/surrealist-memes-regression-progression.}{link}, which serves as an example of how we construct our data in Theperspective.


\begin{table*}[t]\small
    \caption{Data instances in PerSphere. }
    \centering
    \begin{tabular}{p{0.15\textwidth}p{0.22\textwidth}p{0.4\textwidth}p{0.1\textwidth}}
    \toprule
    Query & Claims & Perspectives & Reference \\ \hline
    \hline
    \multirow{4}{=}{\parbox{0.15\textwidth}{Surrealist Memes: Regression or Progression?}} & \multirow{2}{=}{\parbox{0.22\textwidth}{Surrealist memes represent progression in meme art.}} 
    & Surreal memes are derived from a varied line of modern art traditions.  & Doc 205
    \\ \cdashline{3-4}
    && Surrealist memes represent progression in meme art because they demonstrate that no message is necessary for impactful art. & Doc 364
    \\  \cdashline{2-4}

    & \multirow{2}{=}{\parbox{0.22\textwidth}{Surreal art is taking meme art back centuries (well, ok, decade)}} & "Surreal memes are internet elitism at its worst.  & Doc 1138
    \\ \cdashline{3-4}
    && The perspective that surreal art is taking meme art back centuries is supported by the idea that much of its nuance is ``Lost in Translation.''.  & Doc 858
    \\ \midrule

    \multirow{6}{=}{\parbox{0.15\textwidth}{Are Government Soda Taxes Fair?}} & \multirow{2}{=}{\parbox{0.22\textwidth}{The Soda Tax Is a Good Idea}} 
    & The soda tax is a good idea because it can help curb obesity.  & Doc 660
    \\ \cdashline{3-4}
    && The implementation of taxes on cigarettes successfully reduced consumption, suggesting a similar approach with soda could be effective. & Doc 345
    \\  \cdashline{3-4}

    && Investing in America supports the claim that the Soda Tax is a good idea. & Doc 550
    \\  \cdashline{2-4}

    & \multirow{2}{=}{\parbox{0.22\textwidth}{The Soda Tax Is Wrong}} & The Soda Tax is wrong because it infringes on free market principles.  & Doc 769
    \\ \cdashline{3-4}
    && The soda tax is wrong because it only treats a symptom of a larger problem.    & Doc 96 
    \\ \cdashline{3-4}

    &&The inconsistency of food stamp policies, which allow the purchase of sugary drinks, undermines the fairness of the soda tax. & Doc 523

    \\ \midrule

    \multirow{6}{=}{\parbox{0.15\textwidth}{Vaccination must be made compulsory?}} & \multirow{2}{=}{\parbox{0.22\textwidth}{We support the claim Vaccination must be made compulsory.}} 
    & It is the state’s duty to protect its community.  & Doc 3692
    \\ \cdashline{3-4}
    && Compulsory vaccines are a financial relief on the health system. & Doc 8050
    \\  \cdashline{3-4}

    && Duty to protect the child. & Doc 3693, 8047, 2253, 5682, 6052
    \\  \cdashline{2-4}

    & \multirow{2}{=}{\parbox{0.22\textwidth}{We undermine the claim Vaccination must be made compulsory.}} & Compulsory vaccination violates the individuals’ right to bodily integrity.  & Doc 3695, 8052
    \\ \cdashline{3-4}
    && It is a parental right to decide about vaccinations for a child.    & Doc 3696, 3695, 8047 
    \\ \cdashline{3-4}

    &&Vaccines have severe side effects. & Doc 3697, 3469
    \\\hline

    \end{tabular}
    \label{samples}
\end{table*}

\begin{figure*}[p]
    \centering
    \includegraphics[width=0.9\hsize]{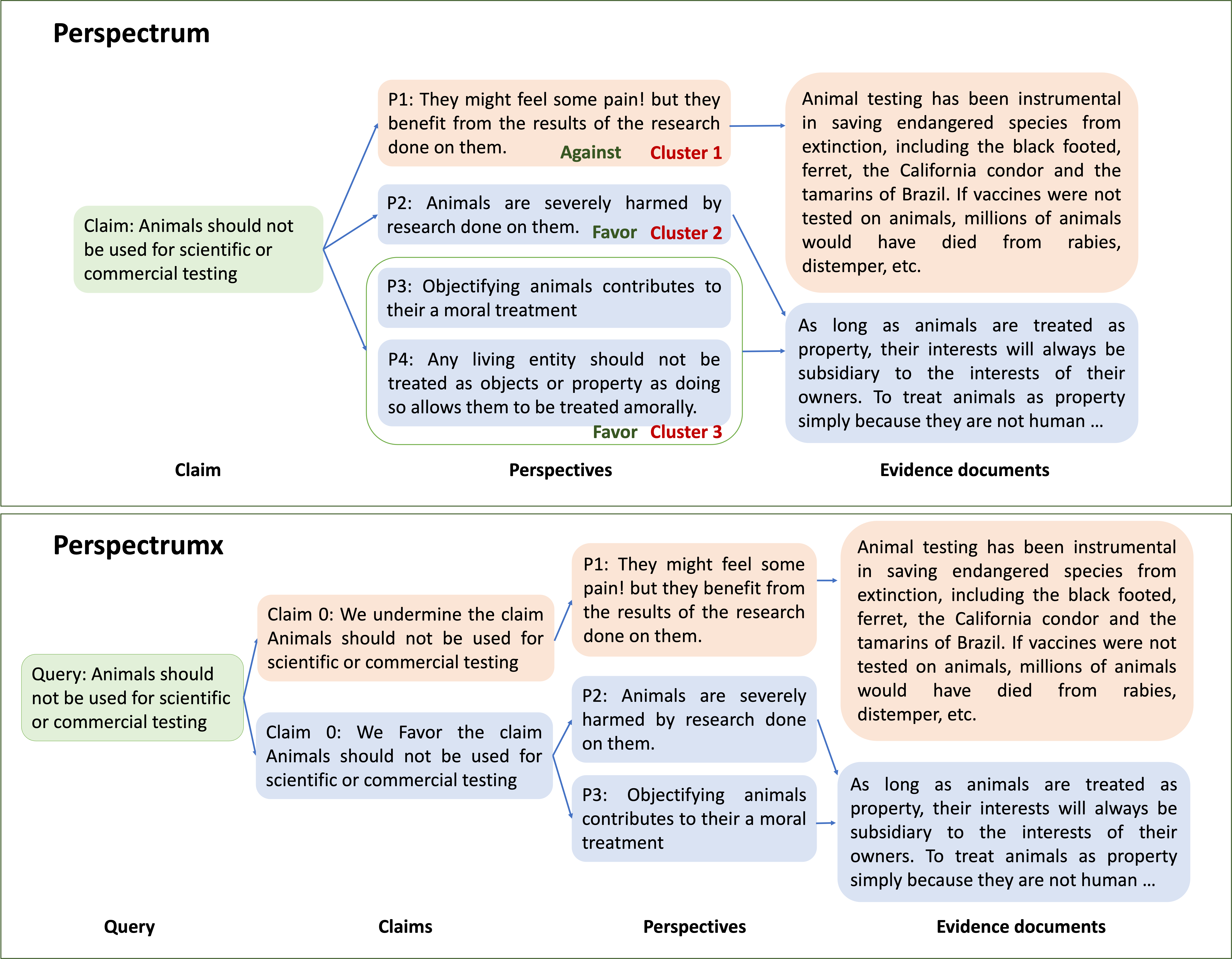}
    \caption{An example of data processing from Perspectrum to Perspectrumx.   
    }
    \label{ins_x}
\end{figure*}

\section{Example for processing Perspectrum}
\label{ins_x_sec}
Figure \ref{ins_x} shows how to process the data in Perspectrum to our data in Perspectrumx.

%% file: acl_latex.bbl
\begin{thebibliography}{39}
\expandafter\ifx\csname natexlab\endcsname\relax\def\natexlab#1{#1}\fi

\bibitem[{Ajith et~al.(2024)Ajith, Xia, Chevalier, Goyal, Chen, and Gao}]{ajith-etal-2024-litsearch}
Anirudh Ajith, Mengzhou Xia, Alexis Chevalier, Tanya Goyal, Danqi Chen, and Tianyu Gao. 2024.
\newblock \href {https://doi.org/10.18653/v1/2024.emnlp-main.840} {{L}it{S}earch: A retrieval benchmark for scientific literature search}.
\newblock In \emph{Proceedings of the 2024 Conference on Empirical Methods in Natural Language Processing}, pages 15068--15083, Miami, Florida, USA. Association for Computational Linguistics.

\bibitem[{Ajjour et~al.(2019)Ajjour, Alshomary, Wachsmuth, and Stein}]{ajjour-etal-2019-modeling}
Yamen Ajjour, Milad Alshomary, Henning Wachsmuth, and Benno Stein. 2019.
\newblock \href {https://doi.org/10.18653/v1/D19-1290} {Modeling frames in argumentation}.
\newblock In \emph{Proceedings of the 2019 Conference on Empirical Methods in Natural Language Processing and the 9th International Joint Conference on Natural Language Processing (EMNLP-IJCNLP)}, pages 2922--2932, Hong Kong, China. Association for Computational Linguistics.

\bibitem[{Chen et~al.(2019)Chen, Khashabi, Yin, Callison-Burch, and Roth}]{chen-etal-2019-seeing}
Sihao Chen, Daniel Khashabi, Wenpeng Yin, Chris Callison-Burch, and Dan Roth. 2019.
\newblock \href {https://doi.org/10.18653/v1/N19-1053} {Seeing things from a different angle:discovering diverse perspectives about claims}.
\newblock In \emph{Proceedings of the 2019 Conference of the North {A}merican Chapter of the Association for Computational Linguistics: Human Language Technologies, Volume 1 (Long and Short Papers)}, pages 542--557, Minneapolis, Minnesota. Association for Computational Linguistics.

\bibitem[{Cinelli et~al.(2021)Cinelli, De~Francisci~Morales, Galeazzi, Quattrociocchi, and Starnini}]{cinelli2021echo}
Matteo Cinelli, Gianmarco De~Francisci~Morales, Alessandro Galeazzi, Walter Quattrociocchi, and Michele Starnini. 2021.
\newblock The echo chamber effect on social media.
\newblock \emph{Proceedings of the National Academy of Sciences}, 118(9):e2023301118.

\bibitem[{Del~Vicario et~al.(2016)Del~Vicario, Vivaldo, Bessi, Zollo, Scala, Caldarelli, and Quattrociocchi}]{del2016echo}
Michela Del~Vicario, Gianna Vivaldo, Alessandro Bessi, Fabiana Zollo, Antonio Scala, Guido Caldarelli, and Walter Quattrociocchi. 2016.
\newblock Echo chambers: Emotional contagion and group polarization on facebook.
\newblock \emph{Scientific reports}, 6(1):37825.

\bibitem[{Edge et~al.(2024)Edge, Trinh, Cheng, Bradley, Chao, Mody, Truitt, and Larson}]{edge2024local}
Darren Edge, Ha~Trinh, Newman Cheng, Joshua Bradley, Alex Chao, Apurva Mody, Steven Truitt, and Jonathan Larson. 2024.
\newblock From local to global: A graph rag approach to query-focused summarization.
\newblock \emph{arXiv preprint arXiv:2404.16130}.

\bibitem[{Ein-Dor et~al.(2020)Ein-Dor, Shnarch, Dankin, Halfon, Sznajder, Gera, Alzate, Gleize, Choshen, Hou et~al.}]{ein2020corpus}
Liat Ein-Dor, Eyal Shnarch, Lena Dankin, Alon Halfon, Benjamin Sznajder, Ariel Gera, Carlos Alzate, Martin Gleize, Leshem Choshen, Yufang Hou, et~al. 2020.
\newblock Corpus wide argument mining—a working solution.
\newblock In \emph{Proceedings of the AAAI Conference on Artificial Intelligence}, volume~34, pages 7683--7691.

\bibitem[{Falkenberg et~al.(2022)Falkenberg, Galeazzi, Torricelli, Di~Marco, Larosa, Sas, Mekacher, Pearce, Zollo, Quattrociocchi et~al.}]{falkenberg2022growing}
Max Falkenberg, Alessandro Galeazzi, Maddalena Torricelli, Niccol{\`o} Di~Marco, Francesca Larosa, Madalina Sas, Amin Mekacher, Warren Pearce, Fabiana Zollo, Walter Quattrociocchi, et~al. 2022.
\newblock Growing polarization around climate change on social media.
\newblock \emph{Nature Climate Change}, 12(12):1114--1121.

\bibitem[{Friedman et~al.(2021)Friedman, Dankin, Hou, Aharonov, Katz, and Slonim}]{friedman-etal-2021-overview}
Roni Friedman, Lena Dankin, Yufang Hou, Ranit Aharonov, Yoav Katz, and Noam Slonim. 2021.
\newblock \href {https://doi.org/10.18653/v1/2021.argmining-1.16} {Overview of the 2021 key point analysis shared task}.
\newblock In \emph{Proceedings of the 8th Workshop on Argument Mining}, pages 154--164, Punta Cana, Dominican Republic. Association for Computational Linguistics.

\bibitem[{Gao et~al.(2023{\natexlab{a}})Gao, Yen, Yu, and Chen}]{gao-etal-2023-enabling}
Tianyu Gao, Howard Yen, Jiatong Yu, and Danqi Chen. 2023{\natexlab{a}}.
\newblock \href {https://doi.org/10.18653/v1/2023.emnlp-main.398} {Enabling large language models to generate text with citations}.
\newblock In \emph{Proceedings of the 2023 Conference on Empirical Methods in Natural Language Processing}, pages 6465--6488, Singapore. Association for Computational Linguistics.

\bibitem[{Gao et~al.(2023{\natexlab{b}})Gao, Xiong, Gao, Jia, Pan, Bi, Dai, Sun, and Wang}]{gao2023retrieval}
Yunfan Gao, Yun Xiong, Xinyu Gao, Kangxiang Jia, Jinliu Pan, Yuxi Bi, Yi~Dai, Jiawei Sun, and Haofen Wang. 2023{\natexlab{b}}.
\newblock Retrieval-augmented generation for large language models: A survey.
\newblock \emph{arXiv preprint arXiv:2312.10997}.

\bibitem[{Glandt et~al.(2021)Glandt, Khanal, Li, Caragea, and Caragea}]{glandt2021stance}
Kyle Glandt, Sarthak Khanal, Yingjie Li, Doina Caragea, and Cornelia Caragea. 2021.
\newblock Stance detection in covid-19 tweets.
\newblock In \emph{Proceedings of the 59th annual meeting of the association for computational linguistics and the 11th international joint conference on natural language processing (long papers)}, volume~1.

\bibitem[{Hu et~al.(2024)Hu, Ru, Qiu, Guo, Zhang, Xu, Luo, Liu, Zhang, and Zhang}]{hu2024refchecker}
Xiangkun Hu, Dongyu Ru, Lin Qiu, Qipeng Guo, Tianhang Zhang, Yang Xu, Yun Luo, Pengfei Liu, Yue Zhang, and Zheng Zhang. 2024.
\newblock Refchecker: Reference-based fine-grained hallucination checker and benchmark for large language models.
\newblock \emph{arXiv preprint arXiv:2405.14486}.

\bibitem[{Huang and Chang(2024)}]{huang2024citation}
Jie Huang and Kevin Chang. 2024.
\newblock Citation: A key to building responsible and accountable large language models.
\newblock In \emph{Findings of the Association for Computational Linguistics: NAACL 2024}, pages 464--473.

\bibitem[{Kamalloo et~al.(2024)Kamalloo, Thakur, Lassance, Ma, Yang, and Lin}]{Kamalloo}
Ehsan Kamalloo, Nandan Thakur, Carlos Lassance, Xueguang Ma, Jheng-Hong Yang, and Jimmy Lin. 2024.
\newblock \href {https://doi.org/10.1145/3626772.3657862} {Resources for brewing beir: Reproducible reference models and statistical analyses}.
\newblock In \emph{Proceedings of the 47th International ACM SIGIR Conference on Research and Development in Information Retrieval}, SIGIR '24, page 1431–1440, New York, NY, USA. Association for Computing Machinery.

\bibitem[{Levy et~al.(2018)Levy, Bogin, Gretz, Aharonov, and Slonim}]{levy-etal-2018-towards}
Ran Levy, Ben Bogin, Shai Gretz, Ranit Aharonov, and Noam Slonim. 2018.
\newblock \href {https://aclanthology.org/C18-1176} {Towards an argumentative content search engine using weak supervision}.
\newblock In \emph{Proceedings of the 27th International Conference on Computational Linguistics}, pages 2066--2081, Santa Fe, New Mexico, USA. Association for Computational Linguistics.

\bibitem[{Li et~al.(2024)Li, Wu, Schlegel, Batista-Navarro, Madusanka, Zahid, Zeng, Wang, He, Li, and Nenadic}]{li-etal-2024-side}
Hao Li, Yuping Wu, Viktor Schlegel, Riza Batista-Navarro, Tharindu Madusanka, Iqra Zahid, Jiayan Zeng, Xiaochi Wang, Xinran He, Yizhi Li, and Goran Nenadic. 2024.
\newblock \href {https://doi.org/10.18653/v1/2024.findings-acl.9} {Which side are you on? a multi-task dataset for end-to-end argument summarisation and evaluation}.
\newblock In \emph{Findings of the Association for Computational Linguistics ACL 2024}, pages 133--150, Bangkok, Thailand and virtual meeting. Association for Computational Linguistics.

\bibitem[{Liu et~al.(2024)Liu, Lin, Hewitt, Paranjape, Bevilacqua, Petroni, and Liang}]{liu2024lost}
Nelson~F Liu, Kevin Lin, John Hewitt, Ashwin Paranjape, Michele Bevilacqua, Fabio Petroni, and Percy Liang. 2024.
\newblock Lost in the middle: How language models use long contexts.
\newblock \emph{Transactions of the Association for Computational Linguistics}, 12:157--173.

\bibitem[{Lo et~al.(2021)Lo, Dai, Xiong, Jiang, and Ku}]{10.1145/3442442.3458613}
Kuan-Chieh Lo, Shih-Chieh Dai, Aiping Xiong, Jing Jiang, and Lun-Wei Ku. 2021.
\newblock \href {https://doi.org/10.1145/3442442.3458613} {Escape from an echo chamber}.
\newblock In \emph{Companion Proceedings of the Web Conference 2021}, WWW '21, page 713–716, New York, NY, USA. Association for Computing Machinery.

\bibitem[{Luo et~al.(2022)Luo, Liu, Shi, Li, and Zhang}]{luo-etal-2022-exploiting}
Yun Luo, Zihan Liu, Yuefeng Shi, Stan~Z. Li, and Yue Zhang. 2022.
\newblock \href {https://aclanthology.org/2022.coling-1.621} {Exploiting sentiment and common sense for zero-shot stance detection}.
\newblock In \emph{Proceedings of the 29th International Conference on Computational Linguistics}, pages 7112--7123, Gyeongju, Republic of Korea. International Committee on Computational Linguistics.

\bibitem[{Muennighoff et~al.()Muennighoff, Hongjin, Wang, Yang, Wei, Yu, Singh, and Kiela}]{muennighoff2024generative}
Niklas Muennighoff, SU~Hongjin, Liang Wang, Nan Yang, Furu Wei, Tao Yu, Amanpreet Singh, and Douwe Kiela.
\newblock Generative representational instruction tuning.
\newblock In \emph{ICLR 2024 Workshop: How Far Are We From AGI}.

\bibitem[{Nguyen(2020)}]{nguyen2020echo}
C~Thi Nguyen. 2020.
\newblock Echo chambers and epistemic bubbles.
\newblock \emph{Episteme}, 17(2):141--161.

\bibitem[{Ni et~al.(2022)Ni, Qu, Lu, Dai, Abrego, Ma, Zhao, Luan, Hall, Chang et~al.}]{ni2022large}
Jianmo Ni, Chen Qu, Jing Lu, Zhuyun Dai, Gustavo~Hernandez Abrego, Ji~Ma, Vincent Zhao, Yi~Luan, Keith Hall, Ming-Wei Chang, et~al. 2022.
\newblock Large dual encoders are generalizable retrievers.
\newblock In \emph{Proceedings of the 2022 Conference on Empirical Methods in Natural Language Processing}, pages 9844--9855.

\bibitem[{Nyhan et~al.(2023)Nyhan, Settle, Thorson, Wojcieszak, Barber{\'a}, Chen, Allcott, Brown, Crespo-Tenorio, Dimmery et~al.}]{nyhan2023like}
Brendan Nyhan, Jaime Settle, Emily Thorson, Magdalena Wojcieszak, Pablo Barber{\'a}, Annie~Y Chen, Hunt Allcott, Taylor Brown, Adriana Crespo-Tenorio, Drew Dimmery, et~al. 2023.
\newblock Like-minded sources on facebook are prevalent but not polarizing.
\newblock \emph{Nature}, 620(7972):137--144.

\bibitem[{Reimer et~al.(2023)Reimer, Bondarenko, Fr{\"o}be, and Hagen}]{reimer-etal-2023-stance}
Jan~Heinrich Reimer, Alexander Bondarenko, Maik Fr{\"o}be, and Matthias Hagen. 2023.
\newblock \href {https://doi.org/10.18653/v1/2023.argmining-1.5} {Stance-aware re-ranking for non-factual comparative queries}.
\newblock In \emph{Proceedings of the 10th Workshop on Argument Mining}, pages 45--51, Singapore. Association for Computational Linguistics.

\bibitem[{Rinott et~al.(2015)Rinott, Dankin, Alzate~Perez, Khapra, Aharoni, and Slonim}]{rinott-etal-2015-show}
Ruty Rinott, Lena Dankin, Carlos Alzate~Perez, Mitesh~M. Khapra, Ehud Aharoni, and Noam Slonim. 2015.
\newblock \href {https://doi.org/10.18653/v1/D15-1050} {Show me your evidence - an automatic method for context dependent evidence detection}.
\newblock In \emph{Proceedings of the 2015 Conference on Empirical Methods in Natural Language Processing}, pages 440--450, Lisbon, Portugal. Association for Computational Linguistics.

\bibitem[{Robertson and Zaragoza(2009)}]{robertsonbm25}
Stephen Robertson and Hugo Zaragoza. 2009.
\newblock \href {https://doi.org/10.1561/1500000019} {The probabilistic relevance framework: Bm25 and beyond}.
\newblock \emph{Found. Trends Inf. Retr.}, 3(4):333–389.

\bibitem[{Sarthi et~al.(2024)Sarthi, Abdullah, Tuli, Khanna, Goldie, and Manning}]{sarthi2024raptor}
Parth Sarthi, Salman Abdullah, Aditi Tuli, Shubh Khanna, Anna Goldie, and Christopher~D Manning. 2024.
\newblock Raptor: Recursive abstractive processing for tree-organized retrieval.
\newblock \emph{arXiv preprint arXiv:2401.18059}.

\bibitem[{See et~al.(2017)See, Liu, and Manning}]{see-etal-2017-get}
Abigail See, Peter~J. Liu, and Christopher~D. Manning. 2017.
\newblock \href {https://doi.org/10.18653/v1/P17-1099} {Get to the point: Summarization with pointer-generator networks}.
\newblock In \emph{Proceedings of the 55th Annual Meeting of the Association for Computational Linguistics (Volume 1: Long Papers)}, pages 1073--1083, Vancouver, Canada. Association for Computational Linguistics.

\bibitem[{Shnarch et~al.(2018)Shnarch, Alzate, Dankin, Gleize, Hou, Choshen, Aharonov, and Slonim}]{shnarch-etal-2018-will}
Eyal Shnarch, Carlos Alzate, Lena Dankin, Martin Gleize, Yufang Hou, Leshem Choshen, Ranit Aharonov, and Noam Slonim. 2018.
\newblock \href {https://doi.org/10.18653/v1/P18-2095} {Will it blend? blending weak and strong labeled data in a neural network for argumentation mining}.
\newblock In \emph{Proceedings of the 56th Annual Meeting of the Association for Computational Linguistics (Volume 2: Short Papers)}, pages 599--605, Melbourne, Australia. Association for Computational Linguistics.

\bibitem[{Stab et~al.(2018)Stab, Daxenberger, Stahlhut, Miller, Schiller, Tauchmann, Eger, and Gurevych}]{stab-etal-2018-argumentext}
Christian Stab, Johannes Daxenberger, Chris Stahlhut, Tristan Miller, Benjamin Schiller, Christopher Tauchmann, Steffen Eger, and Iryna Gurevych. 2018.
\newblock \href {https://doi.org/10.18653/v1/N18-5005} {{A}rgumen{T}ext: Searching for arguments in heterogeneous sources}.
\newblock In \emph{Proceedings of the 2018 Conference of the North {A}merican Chapter of the Association for Computational Linguistics: Demonstrations}, pages 21--25, New Orleans, Louisiana. Association for Computational Linguistics.

\bibitem[{Syed et~al.(2023)Syed, Ziegenbein, Heinisch, Wachsmuth, and Potthast}]{syed-etal-2023-frame}
Shahbaz Syed, Timon Ziegenbein, Philipp Heinisch, Henning Wachsmuth, and Martin Potthast. 2023.
\newblock \href {https://doi.org/10.18653/v1/2023.sigdial-1.10} {Frame-oriented summarization of argumentative discussions}.
\newblock In \emph{Proceedings of the 24th Annual Meeting of the Special Interest Group on Discourse and Dialogue}, pages 114--129, Prague, Czechia. Association for Computational Linguistics.

\bibitem[{Turan et~al.(2019)Turan, Fidan, and Y{\i}ld{\i}ran}]{turan2019critical}
U{\u{g}}ur Turan, Yahya Fidan, and Canan Y{\i}ld{\i}ran. 2019.
\newblock Critical thinking as a qualified decision-making tool.
\newblock \emph{Journal of History, Culture \& Art Research/Tarih K{\"u}lt{\"u}r ve Sanat Arastirmalari Dergisi}, 8(4).

\bibitem[{Van Der~Linden(2022)}]{van2022misinformation}
Sander Van Der~Linden. 2022.
\newblock Misinformation: susceptibility, spread, and interventions to immunize the public.
\newblock \emph{Nature medicine}, 28(3):460--467.

\bibitem[{Wang et~al.(2022)Wang, Yang, Huang, Jiao, Yang, Jiang, Majumder, and Wei}]{wang2022text}
Liang Wang, Nan Yang, Xiaolong Huang, Binxing Jiao, Linjun Yang, Daxin Jiang, Rangan Majumder, and Furu Wei. 2022.
\newblock Text embeddings by weakly-supervised contrastive pre-training.
\newblock \emph{arXiv preprint arXiv:2212.03533}.

\bibitem[{Wang and Ling(2016)}]{wang2016neural}
Lu~Wang and Wang Ling. 2016.
\newblock Neural network-based abstract generation for opinions and arguments.
\newblock In \emph{Proceedings of the 2016 Conference of the North American Chapter of the Association for Computational Linguistics: Human Language Technologies}, pages 47--57.

\bibitem[{Wang et~al.(2024)Wang, Cao, Danek, Zhang, Jin, Lu, and Sun}]{wang2024accelerating}
Zifeng Wang, Lang Cao, Benjamin Danek, Yichi Zhang, Qiao Jin, Zhiyong Lu, and Jimeng Sun. 2024.
\newblock Accelerating clinical evidence synthesis with large language models.
\newblock \emph{arXiv preprint arXiv:2406.17755}.

\bibitem[{Zhang et~al.()Zhang, Kishore, Wu, Weinberger, and Artzi}]{zhangbertscore}
Tianyi Zhang, Varsha Kishore, Felix Wu, Kilian~Q Weinberger, and Yoav Artzi.
\newblock Bertscore: Evaluating text generation with bert.
\newblock In \emph{International Conference on Learning Representations}.

\bibitem[{Zheng et~al.(2023)Zheng, Chiang, Sheng, Zhuang, Wu, Zhuang, Lin, Li, Li, Xing et~al.}]{zheng2023judging}
Lianmin Zheng, Wei-Lin Chiang, Ying Sheng, Siyuan Zhuang, Zhanghao Wu, Yonghao Zhuang, Zi~Lin, Zhuohan Li, Dacheng Li, Eric Xing, et~al. 2023.
\newblock Judging llm-as-a-judge with mt-bench and chatbot arena.
\newblock \emph{Advances in Neural Information Processing Systems}, 36:46595--46623.

\end{thebibliography}
